\documentclass[letterpaper, 10 pt, conference]{ieeeconf}  

\IEEEoverridecommandlockouts                              

\title{\LARGE \bf
Imagine2Act: Leveraging Object-Action Motion Consistency \\
from Imagined Goals for Robotic Manipulation
}

\author{Liang Heng$^{1,2*}$, Jiadong Xu$^{2*}$, Yiwen Wang$^{1,2*}$, Xiaoqi Li$^{1,2* \dagger}$, Muhe Cai$^{1,2}$, \\Yan Shen$^{1,2}$, Juan Zhu$^{2}$, Guanghui Ren$^{3}$, and Hao Dong$^{1,2}$
\thanks{*Equal contribution; $\dagger$ Project Lead; $^{1}$CFCS, School of Computer Science, Peking University;
$^{2}$PrimeBot; $^{3}$AGIBOT}
}

\usepackage{amsmath}
\usepackage{graphicx}
\usepackage{amssymb}
\usepackage{caption}
\usepackage{makecell}
\usepackage{array}
\usepackage{enumerate}
\usepackage{booktabs}
\usepackage{hyperref}
\usepackage{bm}

\begin{document}

\maketitle
\thispagestyle{empty}
\pagestyle{empty}


\begin{abstract}
Relational object rearrangement (ROR) tasks (\emph{e.g. insert flower to vase} ) require a robot to manipulate objects with precise semantic and geometric reasoning. 
Existing approaches either rely on pre-collected demonstrations that struggle to capture complex geometric constraints or generate goal-state observations to capture semantic and geometric knowledge, but fail to explicitly couple object transformation with action prediction, resulting in errors due to generative noise. 
To address these limitations, we propose Imagine2Act, a 3D imitation-learning framework that incorporates semantic and geometric constraints of objects into policy learning to tackle high-precision manipulation tasks. 
We first generate imagined goal images conditioned on language instructions and reconstruct corresponding 3D point clouds to provide robust semantic and geometric priors. 
This imagined goal point clouds serve as additional inputs to the policy model, while an object–action consistency strategy with soft pose supervision explicitly aligns predicted end-effector motion with generated object transformation.
This design enables Imagine2Act to reason about semantic and geometric relationships between objects and predict accurate actions across diverse tasks.
Experiments in both simulation and real world demonstrate that Imagine2Act outperforms previous state-of-the-art policies. Code is fully open-sourced at \url{https://github.com/LiangHeng121/Imagine2Act}.
\end{abstract}


\section{INTRODUCTION}



\begin{figure}[htbp]
    \centering
    \includegraphics[width=0.50\textwidth]{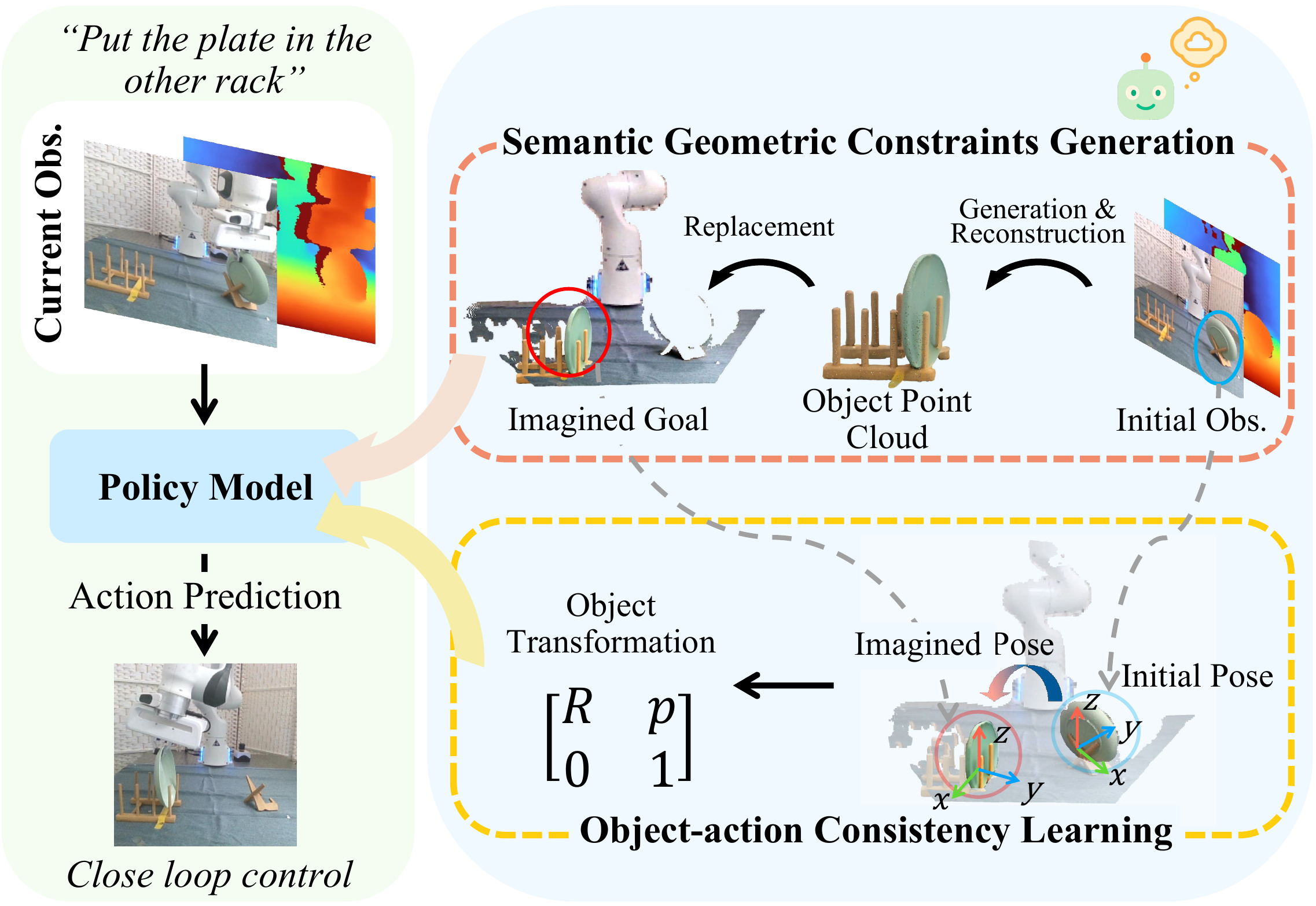}
    \caption{\textbf{Imagine2Act} enhances geometric awareness and improves high-precision tasks' accuracy by introducing semantic geometric constraints generation and object-action consistency learning.}
    \label{fig:teaser}
    \vspace{-0.3cm}
\end{figure}

Relational object rearrangement (ROR)~\cite{batra2020rearrangement, weihs2021visual, li2022stable, yuan2022sornet, goyal2022ifor, chen2022neural, paxton2022predicting} is a fundamental skill for domestic robots, particularly in tasks such as autonomous cleanup and de-cluttering. These tasks require the robot to reason about how objects should be placed and to execute with high precision. Such problems are especially challenging because they involve reasoning over semantic relations between objects as well as handling strict geometric constraints with minimal tolerance.
A representative example is the \textit{Plate-in-rack} task illustrated in Figure~\ref{fig:teaser}, where the policy must adjust the robot's end-effector pose to ensure that the plate is inserted upright into the narrow slot positioned between two adjacent posts of the dish rack.



A common approach in robot learning is 3D imitation learning~\cite{ke20243d, ze20243d, garcia2025towards, goyal2023rvt, gervet2023act3d}, which maps RGB-D observations to robot actions but does not explicitly reason about the complex geometric constraints between objects.
Meanwhile, recent works~\cite{simeonov2023shelving,pan2023tax,chen2022neural,paxton2022predicting} address these tasks by estimating the correspondence between objects to enhance object geometric relationship awareness. 
However, such approaches are limited in that they primarily capture detailed geometric transformations, \emph{without explicitly leveraging the knowledge from the physical world that encodes common-sense semantic constraints between objects}.
For example, in the \textit{Plate-in-rack} task, common sense dictates that the plate should be placed upright between adjacent posts of the rack, rather than incorrectly positioned on top of the posts or laid flat across multiple posts. 
On the other hand, some works~\cite{huang2024imagination,huang2025match} attempt to incorporate common-sense semantic and geometric knowledge by leveraging powerful generative models to produce goal-state observations. 
However, these approaches \emph{fail to explicitly couple action prediction with generated object transformation}.
They either directly execute the generated object transformation, which often fails due to generative inaccuracies, since the generated geometric relationships rarely align exactly with the actual scene, or use the generated goals merely as auxiliary inputs to the policy~\cite{zhong20233d}, without explicitly formulating the correspondence between generated object transformation and end-effector's action motion.

Inspired by these works, we propose \textbf{Imagine2Act}, a 3D imitation-learning framework that \emph{incorporates semantic geometric constraints of objects into policy learning to enhance geometric awareness and enable precise action prediction guided by imagined object transformation signal}.
First, we design a robust \textbf{semantic geometric constraints generation} module that leverages powerful off-the-shelf models to produce an imagined goal, which can generalize across tasks in a zero-shot manner.
Specifically, we use an image-editing model~\cite{openai2025} to generate an imagined goal image depicting the desired semantic and geometric configuration of objects as specified by the language instruction.
We then perform 3D reconstruction~\cite{tochilkin2024triposr} to obtain object point cloud, which are replaced in the scene to form the imagined goal.
This imagined goal point cloud then serves as an additional conditioning input to the policy model, injecting common-sense object relationship knowledge and improving the model’s ability to reason over object semantic geometric constraints.

Although the imagined goal point cloud provides strong geometric constraints guidance to the end-effector motion, directly transferring the object transformation between the initial and imagined goal point clouds to the robot action can fail due to generative noise. 
To address this, we introduce an \textbf{object–action consistency learning} strategy that explicitly couples object transformations with robot actions while avoiding error accumulation. 
Specifically, we estimate the object’s SE(3) transformation between the initial and imagined point clouds and condition the policy model on this object transformation prior.
In addition, we design a soft pose consistency loss on the predicted end-effector actions to align action motion with the generated object transformation under soft supervision, thereby mitigating potential error accumulation.
By doing so, Imagine2Act can reason about the semantic geometric object relationship and predict accurate actions to complete high-precision tasks.

We evaluate Imagine2Act on RLBench~\cite{james2020rlbench} and in the real world. On RLBench across 7 relational object rearrangement tasks, Imagine2Act achieves a mean success rate of 0.79, yielding an absolute improvement of at least 10\% compared to 3D Diffuser Actor~\cite{ke20243d}, Imagine Policy~\cite{huang2024imagination}, and 3D-LOTUS~\cite{garcia2025towards}.
In real-world setting, the policy learns multi-task precise manipulation and delivers consistent improvements across 6 high-precision rearrangement tasks with an average increase of 25\% in success rate compared to 3D Diffuser Actor~\cite{ke20243d}. 
The approach is further applied to articulated object manipulation tasks in RLBench to verify its scalability to other types of tasks, which still shows promising performance.

In summary, our contributions are as follows:
\begin{itemize}
    \item We design an imagined goal point cloud generation module that leverages powerful off-the-shelf models to ensure generation robustness under zero-shot settings. This module provides semantic and geometric object constraints for policy learning.
    \item We design an object–action consistency learning strategy that ensures alignment between predicted end-effector action motion and generated object transformation to effectively leverage semantic geometric prior and avoid error accumulation.
    \item We evaluate on both RLBench and real-world setting, showing consistent gains over strong previous SOTA baselines.
\end{itemize}

\section{RELATED WORK}
\subsection{3D Imitation Learning Policy}
Diffusion policies~\cite{chi2023diffusion} are widely applied in robotics and outperform previous methods such as deterministic behavioral cloning~\cite{florence2022implicit} and Gaussian Mixture Models~\cite{mandlekar2021matters}. However, traditional diffusion policies rely on 2D images and have to learn implicit mappings from 2D to 3D space, leading to camera positioning sensitivity and failure to capture comprehensive spatial information. 
Recent works~\cite{goyal2023rvt, xian2023chaineddiffuser, gervet2023act3d} attempt to fuse 3D scene representations with diffusion policies, achieving significant performance improvements. 
These approaches fall into several paradigms. Multi-view representation methods like RVT~\cite{goyal2023rvt} project 3D point clouds to multiple 2D images, converting the manipulation task into a multi-view policy learning problem. 
Keypose-based approaches like ChainedDiffusor~\cite{xian2023chaineddiffuser}, which employs 3D end-effector keyposes generated from Act3D~\cite{gervet2023act3d} and 3D scene representations to predict end-effector trajectories linking different keyposes.
Building upon these advances, our work introduces semantic and geometric object constraints into policy learning process to enhance geometric reasoning and improve action prediction accuracy.

\subsection{Relational Object Rearrangement from Perception}
Perception-based relational object rearrangement is a crucial problem for manipulation tasks involving object-target interactions (such as stacking and hanging), requiring semantic understanding of physical rules governing geometric relationships. 
End-to-end learning policies~\cite{shridhar2023perceiver,brohan2022rt,ke20243d} struggle to achieve high precision and often fail to generalize across different object categories. 
Previous works on relational object rearrangement primarily focus on configuring keypoints or point cloud representations of objects. Neural Descriptor Fields (NDF)~\cite{simeonov2022neural} encodes points and relative poses using category-level descriptors in a self-supervised manner, but assumes the target is static. 
Relational Pose Diffusion (RPDiff)~\cite{simeonov2023shelving} employs iterative denoising to handle multi-modality scenarios more effectively. 
Imagination Policy~\cite{huang2024imagination} develops a generative point cloud model to predict movement of individual points iteratively, but struggles to guarantee precision during generalization. 
All the aforementioned methods fail to achieve zero-shot object correspondence estimation or goal state generation, whereas our approach leverages the powerful capabilities of foundation models is able to generate imagined goals in a zero-shot manner.
Furthermore, our method introduces an object–action consistency learning strategy to leverage the inherent relationship of object and action motion, enabling robust spatial reasoning across diverse manipulation scenarios and accurate action prediction.

\section{METHOD}

\begin{figure*}[t]
  \centering
\includegraphics[width=0.9\textwidth]{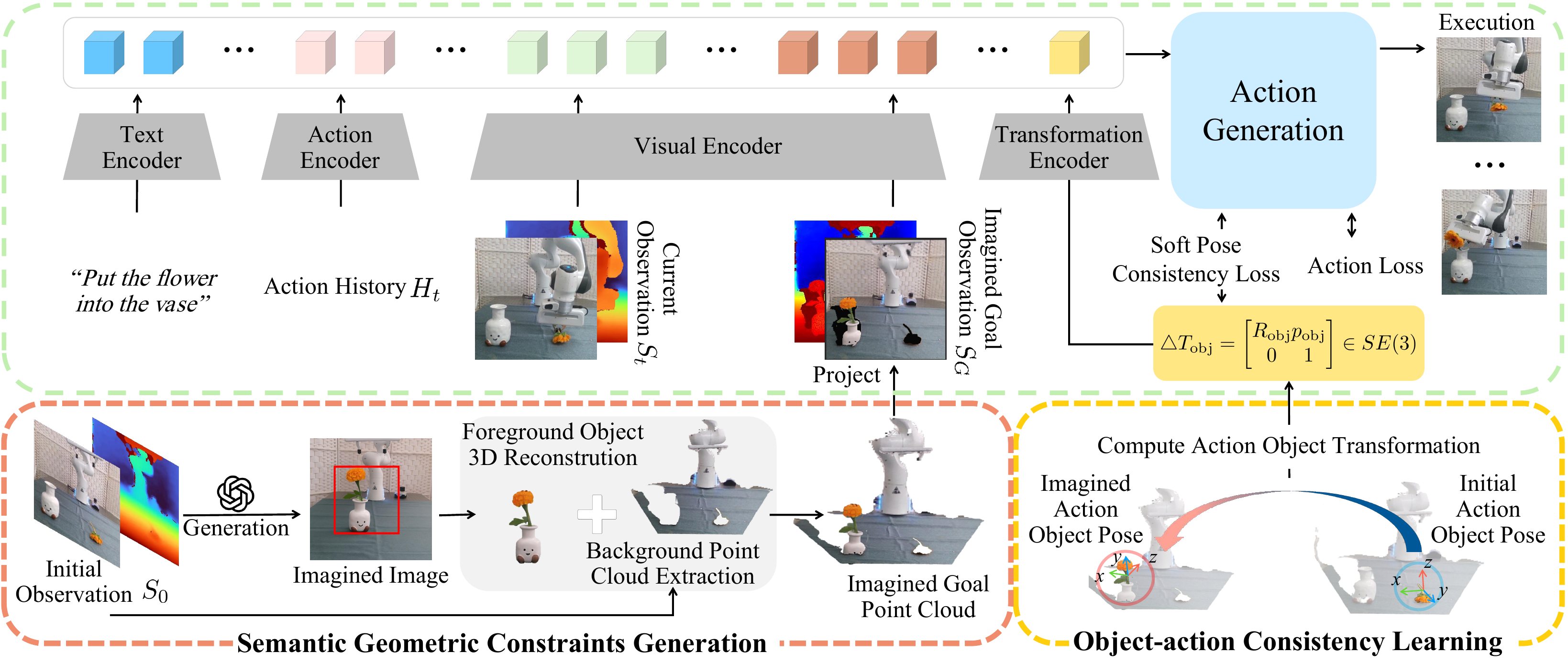}
  \caption{\textbf{Overview of Imagine2Act.} Before robot execution, the semantic–geometric constraint generation module produces an imagined point cloud conditioned on the initial observation. During training, this imagined point cloud is used as an additional input to the policy. Furthermore, by introducing Object–Action Consistency Learning, we compute the transformation between the initial and imagined object poses, which serves as an auxiliary prior input and contributes a loss term that enforces the strong correlation between object transformation and end-effector motion.}
  \label{fig:method}
  \vspace{-0.5cm}
\end{figure*}

\subsection{Problem Statement}
We assume access to a training dataset of $N$ demonstrations of $T$ timesteps:
$\mathcal{D} = \{  \{ \mathcal{S}_{t}^{(i)} \}_{t=0}^{T}, L^{(i)}, \{ a_{t}^{(i)} \}_{t=0}^{T}  \}_{i=1}^{N}$.
At each time step $t$, given a single-view RGB-D scene observation $\mathcal{S}_t$ and a goal $L$ described by a natural language instruction (\emph{e.g.}, ``Put the flower in the vase"), the task is to generate an end-effector action sequence $\hat{\mathcal{A}} = \{\hat{a}_{t+1}, \dots, \hat{a}_{t+k}\}$ of $k$ action chunk size, where $\hat{a}_t \in SE(3)$, which is supervised under groundtruth actions $\mathcal{A} = \{a_{t+1}, \dots, a_{t+k}\}$.

Following TAX-Pose~\cite{pan2023tax}, we define that the observation $S_t$ consists two disjoint sets of objects:
\begin{itemize}
\item Action objects ($O_m$): the objects that the end-effector is expected to manipulate and change their poses during task execution (\emph{e.g.}, a flower in Fig.~\ref{fig:method}).
\item Anchor objects ($O_s$): the objects that should remain fixed and provide semantic geometric constraints for the task (\emph{e.g.}, the vase in Fig.~\ref{fig:method}).
\end{itemize}

The policy model aims to move $O_m$ from its initial pose and gradually adjust it so that the relationship between $O_m$ and $O_s$ satisfies the relational goal specified by $L$. 
Achieving this requires not only predicting feasible actions in SE(3) space, but also reasoning over strict semantic geometric constraints between $O_m$ and $O_s$, since even minor deviations can lead to task failure.

\subsection{Overview and Architecture}
    
At the beginning of an execution process, the proposed \textbf{Semantic Geometric Constraints Generation} (Sec.~\ref{contrib1}) module generates an imagined goal observation $\mathcal{S}_G$ conditioned on the initial observation $\mathcal{S}_0$ and the language instruction $L$. 
This step provides object semantic geometric constraints for the subsequent policy learning, and it is performed before execution without adding extra computational time during execution. 
After that, the diffusion-based policy model $\pi_{\text{action}}$ is then trained to predict action sequence.
During training, we introduce \textbf{Object–action Consistency Learning} (Sec.~\ref{contrib2}) strategy to formulate the relationship between end-effector action motion and generated object transformation. Specifically, we compute the SE(3) transformation of the action object from its initial to its goal state (as generated by the generation module) and encode it into a transformation token. This token is injected into the policy to guide action prediction. 
Furthermore, a soft pose consistency loss is employed to enforce alignment between the predicted action motions and the computed object transformation.

As for the architecture, following 3D Diffuser Actor~\cite{ke20243d}, we adopt a 3D Transformer-based conditional diffusion model. 
Both the current observation $S_{t}$ and the imagined goal observation $S_G$ are processed by a frozen visual encoder (\emph{e.g.}, CLIP~\cite{radford2021learning}) to extract multi-scale semantic tokens, which are unprojected to 3D visual tokens with the depth map.
Language instructions $L$ are projected into the same embedding space by a text encoder (\emph{e.g.}, a pre-trained CLIP language encoder~\cite{gervet2023act3d}) to obtain language tokens. 
We also encode the action history $H_t = \{a_{t-m}, \ldots, a_{t}\}$ ($m$ denoting the history window length) into a set of history state tokens by an action encoder (implemented as MLP), providing temporal context. 
All aforementioned tokens, including a transformation token, which will be illustrated in detail in Sec.~\ref{contrib2}, are concatenated and processed by a diffusion transformer~\cite{ke20243d} for action generation.  

\subsection{Semantic Geometric Constraints Generation} 
\label{contrib1}

In this section, we describe how we obtain robust semantic geometric constraints of objects, which then serve as input conditions for policy learning to enhance geometric awareness. 
The key idea is to leverage powerful external models to generate imagined goal observations that can generalize across tasks and scenes in a zero-shot manner, while minimizing potential noise by editing only the relevant objects and keeping the rest of the scene unchanged.
Concretely, this module consists of two stages: generating the goal observation and conditioning the policy model on it.

\subsubsection{Generating Goal Observation}





Our goal is to construct a robust imagined 3D goal observation that encodes the semantic and geometric constraints of objects, serving as a conditioning input for policy learning. To achieve this, rather than directly generating 3D goals, we decompose the process into two steps. First, we leverage large generative models to produce an imagined image, as these models are trained on large-scale image datasets and can capture semantic layouts and object relationships with high fidelity. Then, conditioned on the imagined goal image, we reconstruct the corresponding imagined goal point cloud while striving to preserve accuracy and consistency with the actual scene.

Concretely, to harness the strong semantic reasoning capabilities of large generative models, we first generate an imagined image depicting the scene at task completion, conditioned on the initial observation $S_0$ and the language instruction $L$, using a generative model (e.g., GPT-Image-1~\cite{openai2025}).
To ensure consistency with the actual scene, the generated image is constrained to match the camera view of the initial observation. This alignment is critical because the subsequent 3D reconstruction assumes the same viewpoint for proper spatial alignment with the initial 3D point cloud.

Next, we aim to construct the corresponding imagined goal point cloud. To minimize generative noise, we ensure that only task-relevant objects are modified while keeping the rest of the scene unchanged. Specifically, we first segment the foreground objects from the imagined image using a segmentation model (e.g., Grounded-SAM~\cite{ren2024grounded}) guided by the instruction, separating the imagined final states of foreground objects (action object $O_m$ and the anchor object $O_s$) from the background. The background point cloud $P_{\text{back}}$ is directly extracted from the initial observation, while the foreground objects are reconstructed from the imagined image using a 3D reconstruction model (e.g., TripoSR~\cite{tochilkin2024triposr}) to obtain their point clouds $P_{fore}$, which encode the imagined geometric constraints.


We observe that the pose of the anchor object is usually unchanged throughout the manipulation process, enabling us to place the foreground objects back in the scene in the anchor object's pose and scale.
For pose determination, the anchor object $O_s$'s 6D pose $T_{anchor}^{pose} \in \operatorname{SE}(3)$ is estimated from the initial observation $S_0$ and camera parameters using a 6D pose estimation algorithm (e.g., FoundationPose~\cite{wen2024foundationpose}). 
This anchor allows proper alignment of the generated foreground objects with the background scene in the world coordinate system. 
For scale determination, a factor $s$ is manually set to ensure that the reconstructed object dimensions match the real-world scale observed in $S_0$, and this scale is applied uniformly across tasks via a scaling matrix $T_{anchor}^{scale} = \text{diag}(s,s,s,1)$.
Finally, the imagined goal point cloud is assembled as:
\begin{equation}
P_G = P_{\text{back}}\cup \big(T_{anchor}^{pose} \cdot T_{anchor}^{scale} \cdot P_{fore}\big)
\end{equation}
, where ``$\cdot$" denotes applying the rigid transform to each point cloud. 
This generation pipeline ensures that the imagined goal point cloud accurately encodes semantic and geometric constraints while remaining aligned with the actual scene for downstream policy learning.

\subsubsection{Conditioning as Policy Input}
To ensure that the imagined goal observation $S_G$ can be seamlessly integrated into the imitation learning policy, which requires RGB-D inputs, we project the assembled imagined goal point cloud $P_G$ to obtain the corresponding $S_G$ RGB and depth maps. We then process $S_G$ in the same manner as the current observation $S_t$ for feature extraction.
Specifically, at each time step $t$, the visual input of the model consists of two parts:

\begin{enumerate}[a)]
    \item \textbf{Current observation}: the RGB-D data $S_t$ captured at timestep $t$.
    \item \textbf{Imagined goal observation}:  the imagined goal RGB-D observation $S_G$, which is obtained before robot execution.
\end{enumerate}

We use a shared visual encoder to process both the initial and goal observations, extracting 3D visual tokens for predicting accurate actions conditioned on the underlying semantic and geometric constraints.

\begin{table*}[t]
\centering
\renewcommand{\arraystretch}{1.2}
\setlength{\tabcolsep}{4pt}
{\normalsize
\caption{\textbf{Evaluation in RLBench of relational object rearrangement tasks.} We report the success rate across 7 tasks. Visualization is shown on the left side of Figure.~\ref{fig:sim}.
The last column reports the margin of Imagine2Act over each baseline.}
\label{tab:sim}
\resizebox{\textwidth}{!}{
\begin{tabular}{l|ccccccc|cc}
\toprule
Method & Phone & Put-Knife & Stack-Wine & Put-Plate & Put-Roll & Stack-Cups & Place-Cups & Avg & Margin \\
\midrule
3DDA & \textbf{1.00} & 0.28 & 0.92 & 0.84 & 0.44 & 0.84 & 0.40 & 0.67 & +0.12 \\
Imagine Policy & 0.88 & 0.24 & 0.60 & 0.28 & 0.16 & 0.12 & 0.12 & 0.34 & +0.45 \\
3D-LOTUS & \textbf{1.00} & \textbf{0.48} & 0.76 & 0.76 & 0.36 & 0.92 & \textbf{0.56} & 0.69 & +0.10 \\\midrule
\textbf{Imagine2Act (Ours)} & \textbf{1.00} & \textbf{0.48} & \textbf{1.00} & \textbf{1.00} & \textbf{0.48} & \textbf{1.00} & \textbf{0.56} & \textbf{0.79} & -- \\
\bottomrule
\end{tabular}
}}
\vspace{-0.6cm}
\end{table*}

\subsection{Object-Action Consistency Learning} 
\label{contrib2}
With the imagined goal observation $S_G$, we are able to compute the rigid-body transformation required to move the movable object $O_m$ from its initial pose to the imagined goal pose.
Since the end-effector is the direct actuator of object motion, its trajectory inherently shows similarity with the object’s transformation, making the two strongly correlated
However, directly using generated object motion as end-effector's action motion may lead to error accumulation due to potential errors in the generation process. 
Based on this observation, we propose exploiting the strong correlation between object transformation and end-effector'action motion: on the one hand, we encode the SE(3) transformation of action object $O_m$ as a \emph{transformation token} and inject it into the policy; on the other hand, we introduce a \emph{soft pose-consistency loss} to constrain the predicted actions to remain aligned with the object SE(3) transformation while avoiding error accumulation.


\subsubsection{Encoding Transformation Token}
In this section, we aim to estimate the SE(3) transformation of the action object from its initial pose to the imagined goal pose. This transformation is then encoded into a compact representation, the transformation token, which is injected into the policy model.

Specifically, we first extract the point cloud of the action object from both the initial observation and the imagined goal observation based on the RGB-D frames. Subsequently, we apply a rigid registration procedure (e.g., Kabsch algorithm~\cite{kabsch1976solution,kabsch1978discussion}) between the initial and imagined goal point cloud to compute the object transformation:
\begin{equation}
    T_{\text{obj}} = 
    \begin{bmatrix}
    R_{\text{obj}} & p_{\text{obj}} \\
    0 & 1
    \end{bmatrix}
    \in SE(3)
\end{equation}
where $R_{\text{obj}} \in \operatorname{SO}(3)$ and $p_{\text{obj}} \in \mathbb{R}^3$ denote the optimal rotation matrix and translation vector aligning the two point sets.


To inject this object-level motion into the policy, we use a transformation encoder to encode $T_{\text{obj}}$ into a \emph{transformation token} $\bm{\tau}_{\text{T}}$. Concretely, the rotation matrix $R_{\text{obj}}$ and translation vector $p_{\text{obj}}$ are first flattened into a 12-dimensional vector.
After that, this vector is processed by a scalar encoder followed by a length-aggregation module (both of them implemented as MLP) to yield a single token $\bm{\tau}_{\text{T}}$, which represents the intended transformation of the object. 
Finally, $\bm{\tau}_{\text{T}}$ is concatenated with language tokens, visual tokens, and history state tokens, serving as the input to the action generation module.

\subsubsection{Soft Pose Consistency Loss}
We introduce a soft pose consistency loss, which penalizes deviations between the predicted end-effector motion and the action object's transformation $T_{obj}$, with a threshold to prevent errors in the object transformation from adversely affecting the action prediction.

This loss only applies when the robot has already grasped the action object. Let the predicted poses of the end-effector at the grasp stage and current stage of the manipulation phase be \(\hat{a}_{g}\) and \(\hat{a}_{t}\), respectively. The predicted relative transformation is then calculated as:

\begin{equation}
\hat{T}_{\text{act}} = \hat{a}_{t} \cdot \hat{a}_{g}^{-1} =
\begin{bmatrix}
\hat{R}_{\text{act}} & \hat{p}_{\text{act}} \\
0 & 1
\end{bmatrix}
\in SE(3)
\end{equation}

Instead of enforcing a strict L2 penalty, which could over-constrain the policy and amplify error accumulation, we adopt a flexible, threshold-based loss function. 
This design offers greater tolerance to small deviations that may not affect task success and potential estimation errors in the computed object transformation, thereby improving both training stability and robustness. 
This soft consistency loss penalizes deviations only when predicted action motion $\hat{T}_{\text{act}}$ exceeds tolerance thresholds relative to the object transformation $T_{\text{obj}}$.

The loss is composed of two terms: a rotation component and a translation component. The rotational deviation is measured by the geodesic distance \(\theta\) between the action and object rotations:
\begin{equation}
    \theta = \arccos\left(\frac{\text{Tr}(\hat{R}_{\text{act}}^TR_{\text{obj}})-1}{2}\right)
\end{equation},
where $\text{TR}(\cdot)$ denotes the matrix trace.

The translation deviation \( d \) is measured by the Euclidean distance between the action and object translations:
\begin{equation}
    d = \| \hat{p}_{\text{act}} - p_{\text{obj}} \|_2
\end{equation}
The soft consistency loss, \( \mathcal{L}_{\text{soft}} \), is then formulated using the sigmoid function \( \sigma(\cdot) \) to create smooth penalties that activate when errors surpass the thresholds \( \tau_{\text{r}} \) and \( \tau_{\text{t}} \):
\begin{equation}
    \mathcal{L}_{\text{r}} = \sigma(k_{\text{r}} \cdot (\theta - \tau_{\text{r}}))
\end{equation}
\begin{equation}
\mathcal{L}_{\text{t}} = \sigma(k_{\text{t}} \cdot (d - \tau_{\text{t}}))
\end{equation}
\begin{equation}
\mathcal{L}_{\text{soft}} = \mathbb{E}[\mathcal{L}_{\text{r}} + \mathcal{L}_{\text{t}}]
\end{equation}

where \( k_{\text{r}} \) and \( k_{\text{t}} \) are scaling factors. In our implementation, we set the tolerance thresholds to \( \tau_{\text{r}} \approx 0.1 \) radians and \( \tau_{\text{t}} = 0.01 \) meters.

\subsection{Objective Function}
The model is trained with a combined objective function consisting of a standard action prediction loss and our proposed soft pose consistency loss. 
During training, we follow the standard~\cite{ke20243d} denoising diffusion probabilistic model objective $\mathcal{L}_{\text{diff}}$. 
Given a ground-truth action sequence, we add Gaussian noise at a random diffusion timestep to obtain a perturbed sequence. The action generation module is trained to predict the added noise.
The final training objective combines the action prediction loss with the soft pose consistency loss:
\begin{equation}
    \mathcal{L} = \mathcal{L}_{\text{diff}} + \lambda_{\text{pose}} \, \mathcal{L}_{\text{soft}}
\end{equation}
, where $\lambda_{\text{pose}}$ is a weighting coefficient.

\begin{figure*}[th] 
    \centering
    \includegraphics[width=0.85\linewidth]{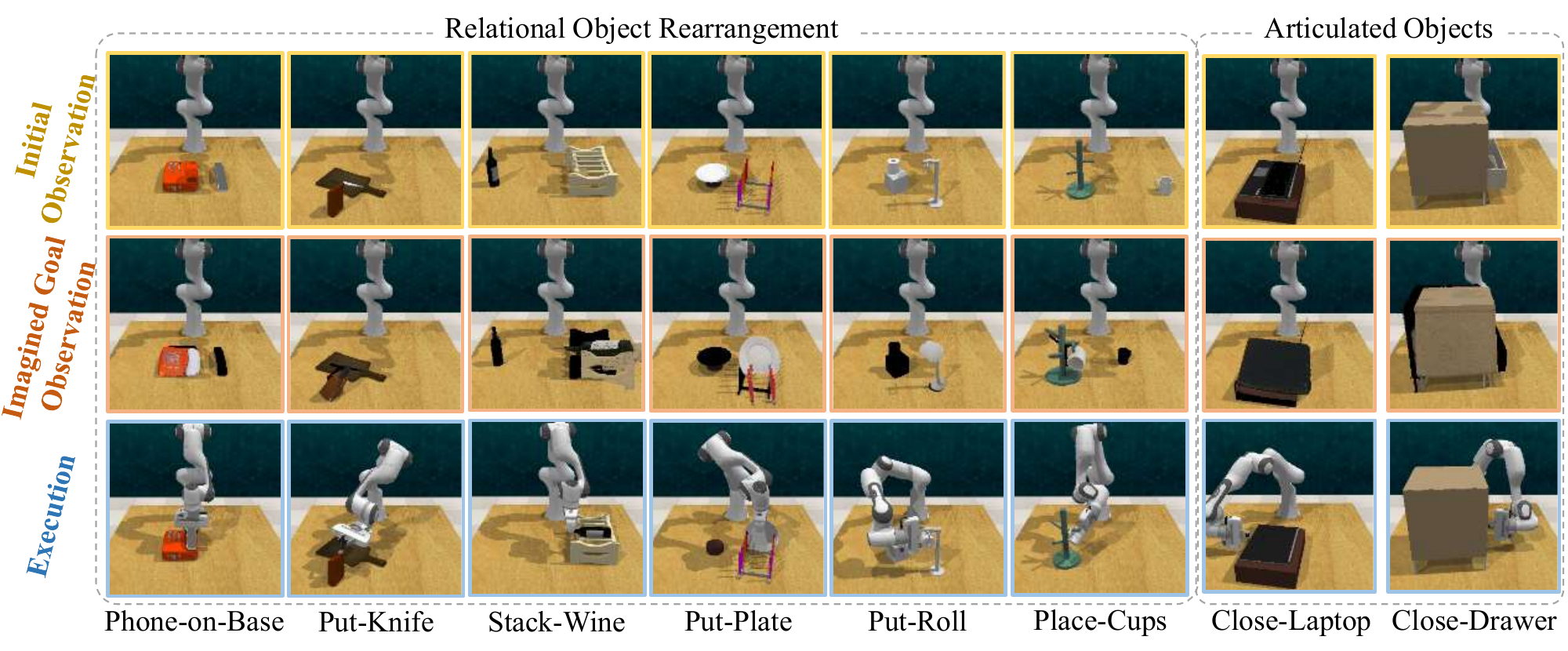}
    \caption{\textbf{Visualization of RLBench Experiments.} We visualize the initial observation, imagined goal point cloud, and execution results of 6 relational object rearrangement tasks and 2 articulated object manipulation tasks.}
    \label{fig:sim}
\vspace{-0.6cm}
\end{figure*}

\section{EXPERIMENTS}
\subsection{Evaluation on RLBench}
We evaluate our method on RLBench. RLBench\cite{james2020rlbench} is a large-scale benchmark and simulation suite for robotic manipulation, built on top of CoppeliaSim\cite{rohmer2013vrep}.

\textbf{Settings.}
All methods, including baselines, are trained on multi-task setup. We select 7 representative relational object rearrangement tasks: \textit{Phone-on-Base}, \textit{Put-Knife}, \textit{Stack-Wine}, \textit{Put-Plate}, \textit{Put-Roll}, \textit{Stack-Cups}, \textit{Place-Cups}, which require high-precision manipulation.
To make a fair comparison, all methods are trained on the single-view setup, using only the front RGB-D camera. 
Following 3D diffuser actor~\cite{ke20243d}, for each task, we use 100 demonstrations for training and evaluate on 25 trials. 
Performance is evaluated by average success rate of each task, and we also calculate the mean success rate averaged over 7 tasks for each method.

\textbf{Baselines}
We compare our method against: \textbf{\textit{3D Diffuser Actor}}\cite{ke20243d} (3DDA) is a typical conditional diffusion model that combines 3D scene representations and diffusion objectives. 
\textbf{\textit{Imagine Policy}}\cite{huang2024imagination} is tailored for representative relational object tasks. It generates a goal point cloud to imagine the goal object state and then uses the generated object transformation directly as actions. 
\textbf{\textit{3D-LOTUS}}\cite{garcia2025towards} is a SOTA 3D robotic manipulation policy that leverages language-conditioned point cloud transformers for action prediction, achieving strong efficiency and performance on tasks. 
\begin{table}[t]
\centering
\renewcommand{\arraystretch}{1.0}
\setlength{\tabcolsep}{2pt}
\vspace{-0.0cm}
\caption{\textbf{Ablation study on Imagine2Act.} Each variant selectively removes or changes components to assess their contributions.}
\label{tab:ablation}
\resizebox{\linewidth}{!}{
\begin{tabular}{l|cc|cc|c}
\toprule
 & \makecell{Transformation\\Token} & \makecell{Soft\\Pose Loss} & \makecell{Imagined\\Point Cloud} & \makecell{GT Point\\Cloud} & \normalsize Avg. \\
\midrule
Ex0 & - & - & - & - & 0.67 \\\midrule
Ex1 & - & - & - & \checkmark & 0.74 \\
Ex2 & - &-  & \checkmark &-  & 0.72 \\\midrule
Ex3 & \checkmark & - & \checkmark &-  & 0.76 \\
Ex4 &-  & \checkmark & \checkmark &-  & 0.75 \\\midrule
Ex5 & \checkmark & \checkmark & \checkmark & - & 0.79 \\
\bottomrule
\end{tabular}
}
\vspace{-0.6cm}
\end{table}

\textbf{Results}
The results of all methods are summarized in Table~\ref{tab:sim}, where Imagine2Act outperforms all baselines across the seven tasks, achieving an average success rate of 0.79. 
Imagine2Act demonstrates particularly strong performance on tasks that require precise semantic and geometric constraints reasoning, such as Put-Knife, where baseline methods often struggle. 
Compared to 3DDA and 3D-Lotus, which directly map 3D observations to actions, Imagine2Act surpasses them by 0.12 and 0.10, respectively, highlighting the benefit of incorporating semantic geometric object constraints into policy learning. 
In comparison with the Imagine policy, which also generates semantic geometric object constraints, Imagine2Act achieves superior performance by avoiding the direct use of generated object transformation as actions, thereby mitigating the impact of generative noise. 
Instead, our method treats these constraints as a policy prior and introduces soft supervision, which formulates the relationship between object transformation and action motion and prevents error accumulation.
We visualize the initial observation, the imagined goal point cloud, and the execution results on the left side of Figure~\ref{fig:sim}.

\begin{figure*}[t] 
    \centering
    \includegraphics[width=0.85\linewidth]{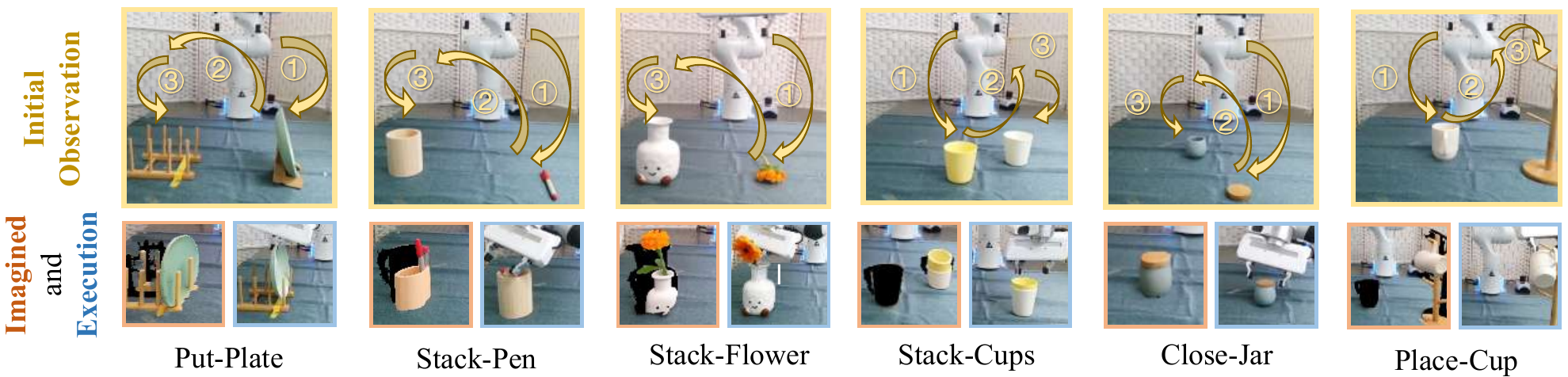}
     \captionsetup{width=1\textwidth}
    \caption{\textbf{Visualization of Real-world Experiment.} We visualize the initial observation and illustrate the required trajectory with yellow arrows. We also display the imagined goal point cloud in the real world and present the manipulation result.}
    \label{fig:real}
\vspace{-0.3cm}
\end{figure*}
\begin{table*}[t]
\centering
\renewcommand{\arraystretch}{0.9}
\setlength{\tabcolsep}{8pt}
\caption{\textbf{Evaluation in Real-world.} Success rate is reported as the number of successes out of 10 trials.}
\label{tab:real}
\resizebox{\textwidth}{!}{
\begin{tabular}{l|cccccc|c}
\toprule
Method & Put-Plate &  Stack-Pen & Stack-Flower & Stack-Cups & Close-Jar  & Place-Cup & Avg.\\
\midrule
3DDA & 6/10 & 6/10 & 6/10 & 3/10 & 2/10 & 3/10 & 0.43\\
\textbf{Imagine2Act} & \textbf{9/10} & \textbf{8/10} & \textbf{8/10} & \textbf{6/10} & \textbf{5/10} & \textbf{5/10} & \textbf{0.68}\\
\bottomrule
\end{tabular}
}
\vspace{-0.6cm}
\end{table*}
\subsection{Ablation Study}

To validate the effectiveness of each module, we construct the following ablated configurations in Table~\ref{tab:ablation}.

\textit{Ex0 Imagine2Act w/o imagine.}
 This variant removes the goal observation generation module, forcing the policy to directly predict actions solely based on the current observation.

 \textit{Ex1 Imagine2Act w/ gt goal.}
This configuration serves as an upper bound setting. Compared with Ex0, it augments the policy input with ground-truth goal observations, which is the final state in the collected demonstration. 
Such information can be obtained in the simulator, since all test trials have pre-collected trajectories.
However, since such supervision is not available in practice, this setting is used only to illustrate how closely the imagined goal observation in Ex2 can approximate the effect of having perfect goal supervision, aiming to reflect the effectiveness of constraints generation module.
 
 \textit{Ex2 Imagine2Act w/ imagine only.}
It incorporates semantic geometric constraints generation module to generate imagined goal observation. However, it simply serves as input to the policy model without applying object-action consistent learning strategy to guide the policy learning.


\textit{Ex3 Imagine2Act w/o soft loss.}
Compared with Ex2, this configuration further introduces the transformation token as an input prior to the policy, enabling the model to leverage object transformation knowledge.  
However, it lacks the soft pose-consistency loss $\mathcal{L}_{\text{soft}}$ to penalize the deviations of predicted action and generated object transformation.

\textit{Ex4 Imagine2Act w/o transformation token.}
This variant builds upon Ex2 by adding the soft pose-consistency loss $\mathcal{L}_{\text{soft}}$, which regularizes the alignment between predicted action and generated object transformations. 
However, it lacks the transformation token input, so the policy is not explicitly conditioned on object transformations.

We report the ablation results in Table~\ref{tab:ablation}. 
Comparing Ex0 and Ex1\&2, removing the imagination module leads to a large performance drop, demonstrating that incorporating semantic geometric object constraint is crucial for policy learning. 
Moreover, Ex2, which uses the generation manner to obtain goal state, achieves similar performance compared to Ex1, which uses the actual goal state observation ($S_{T}$) as the model input.
This indicates that our generation module is robust and accurate enough to ensure a high-quality goal point cloud generation under a zero-shot manner.
Comparing Ex2 and Ex3, adding the transformation token leads to improved performance, confirming that explicitly conditioning the policy on object transformations helps capture object motion prior effectively.
Similarly, adding the soft pose-consistency loss (Ex4) on top of Ex2 also improves success rates, demonstrating that the soft loss is effective in aligning predicted action motions with imagined object transformations.
Finally, combining all components (Ex5) achieves the best overall performance, showing that all the proposed components work jointly to realize the effective policy model learning.

\subsection{Evaluation in Real World}
We validate the proposed Imagine2Act on a real robot and use 3D Diffuser Actor (3DDA) as a baseline for comparison. We select it as the baseline because it achieves relatively strong performance in the simulator experiments, and since our method shares the same policy model architecture as 3DDA, comparing with it provides the most direct insight into the effectiveness of our proposed module.

\textbf{Settings}
The real-world experiment is performed on a Franka Emika robot equipped with an RGB-D RealSense 435 camera at a front view.
We evaluate our method and 3D Diffuser Actor on 6 manipulation tasks. 
\textbf{\textit{Stack-Cups}}: The robot is asked to pick up one cup and place it into another cup. \textbf{\textit{Close-Jar}}: This task consists of grasping the lid and then placing it on the jar to seal it. \textbf{\textit{Stack-Flower}}: The agent must pick up a flower, position its stem vertically, and insert it into a vase. \textbf{\textit{Stack-Pen}}: The agent needs to pick up a pen and insert it into a pen holder.  \textbf{\textit{Put-Plate}}: The robot should grasp a plate from the table and place it upright on a designated plate rack. \textbf{\textit{Place-Cup}}: The robot needs to grab a cup from the table and hang it on a designated cup holder. For each task, we collect 50 demonstrations for training and 10 demonstrations for validation. We use the success rate of 10 demonstrations as our evaluation metric.

\textbf{Results}
As shown in Table~\ref{tab:real}, Imagine2Act consistently outperforms 3D Diffuser Actor across all tasks. On Stack-Cups and Close-Jar, Imagine2Act achieves 6/10 and 5/10 successes respectively, achieving almost twice the success rate of the baseline. In stack-pen and put-plate, our method shows strong reliability with 8/10 and 9/10 success rates, significantly higher than the baseline's 6/10. 
These results highlight the robustness of our method in real-world manipulation scenarios.
In Figure~\ref{fig:real}, for each task, we visualize the initial observation, the imagined goal point cloud, and the final state after execution.

\subsection{Articulated Object Manipulation}
\begin{table}[t]
\centering
\renewcommand{\arraystretch}{1.0}
\setlength{\tabcolsep}{3pt}
{\normalsize
\caption{\textbf{Evaluation in RLBench of articulated object manipulation tasks.} We report the success rate across 5 tasks.}
\label{tab:articulated}
\resizebox{0.5\textwidth}{!}{
\begin{tabular}{l|ccccc|c}
\toprule
Method & \makecell{Close-\\Box} & \makecell{Close-\\Laptop} & \makecell{Close-\\Drawer} & \makecell{Open-\\Microwave} & \makecell{Close-\\Fridge} & Avg.\\
\midrule
3DDA & 0.96 & \textbf{1.0} & 0.92 & 0.84 & \textbf{1.0} & 0.94\\
\makecell{\textbf{Imagine2Act}} & \textbf{1.0} & \textbf{1.0} & \textbf{0.96} & \textbf{0.92} & \textbf{1.0} & \textbf{0.98}\\
\bottomrule
\end{tabular}
}}
\vspace{-0.6cm}
\end{table}
Articulated object manipulation poses unique challenges due to the need to reason over object kinematics, such as revolute and prismatic joints. 
We further validate our method on these tasks in RLBench simulator to demonstrate the effectiveness of Imagine2Act beyond relational object rearrangement tasks.
Specifically, we include five representative tasks (Figure~\ref{fig:sim}) : \textbf{\textit{close-box}}, \textbf{\textit{close-laptop-lid}}, \textbf{\textit{close-drawer}}, \textbf{\textit{close-fridge}}, and \textbf{\textit{open-microwave}}. 
We handle articulated objects by segmenting the action and anchor object parts, and fuse them into a unified representation, enabling the policy to reason about joint kinematics.
As shown in Table~\ref{tab:articulated}, our method achieves comparable performance to 3D Diffuser Actor across these articulated object manipulation tasks, demonstrating that the proposed approach can generalize to other types of manipulation. 
We visualize two tasks on the right side of Figure~\ref{fig:sim}, showing the effectiveness of the proposed method on articulated object manipulation.
Note that our model architecture is identical to that of 3D Diffuser Actor; since 3DDA already performs well on these tasks, the margin between the two methods is small. However, for relational object manipulation tasks with high-precision requirements, our method shows significant advantages.


\section{CONCLUSION}

We introduce Imagine2Act, a 3D imitation-learning framework for tasks requiring precise semantic and geometric reasoning. 
By generating imagined goal point cloud, our method provides robust semantic and geometric priors to the policy. 
The designed object–action consistency strategy with soft pose supervision aligns predicted actions with object transformations, enabling accurate, high-precision manipulation. Experiments in simulation and the real world demonstrate that Imagine2Act outperforms prior state-of-the-art policies across diverse tasks.












{
\bibliographystyle{IEEEtran}
\bibliography{root}
}

\end{document}